\documentclass[11pt]{article}

\usepackage[preprint]{acl}

\usepackage{times}
\usepackage{latexsym}
\usepackage[T1]{fontenc}
\usepackage[utf8]{inputenc}
\usepackage{microtype}
\usepackage{inconsolata}
\usepackage{graphicx}
\usepackage{booktabs}
\usepackage{amsmath}
\usepackage{amssymb}
\usepackage{listings}
\usepackage{tabularx}
\usepackage{array}
\lstdefinestyle{tsearchprotocol}{
  basicstyle=\ttfamily\normalsize,
  columns=fullflexible,
  keepspaces=true,
  showstringspaces=false,
  breaklines=true,
  breakatwhitespace=true,
  breakautoindent=false,
  breakindent=0pt,
  emptylines=0,
  aboveskip=6pt,
  belowskip=6pt,
  escapeinside={(*@}{@*)}
}
\graphicspath{{figures/}}

\title{T-Search: An Open Agentic Retriever and Playground\\for Hard Multi-Step Search}

\author{\textbf{Olga Tsymboi}, \textbf{Ramil Latypov}, \textbf{Aleksandr Medvedev},
  \textbf{Danil Taranets},
  \textbf{Dmitrii Stoianov}, \\ \textbf{Nikita Gulyakov}, \textbf{Gleb Alektorov}, \textbf{Anatolii Potapov} \\
  T-Tech \\
  \small{
   \textbf{Correspondence:} \href{mailto:anatolii.s.potapov@gmail.com}{anatolii.s.potapov@gmail.com}
 }
}

\begin{document}
\maketitle
\begin{abstract}
We present T-Search, an open-weight agentic retriever for hard multi-step search. Given a question and a search tool over a fixed corpus, it runs a bounded multi-round search and returns a ranked list of evidence chunks with short justifications, leaving answer generation to a downstream model, so backend and generator can be swapped without retraining. T-Search is built on Qwen3.6-35B-A3B and trained on adversarially filtered synthetic search tasks with round-sliced supervised fine-tuning followed by GSPO on a recall reward. Averaged over seven English and Russian benchmarks with gold evidence annotations, it reaches 56.0 Recall@10 with one rollout, 14.4 points above its base, and 61.3 with three fused rollouts, outperforming larger open models. We release the model, harness, live demo, and three benchmarks, including TRuST, the first native-Russian hard-search benchmark.
\end{abstract}

\section{Introduction}
\label{sec:intro}

Retrieval-augmented generation \citep{lewis2020rag} assumes that the answer can be found by querying the question as posed. Multi-hop questions broke this assumption with HotpotQA \citep{yang2018hotpotqa}, and iterative retrieve-and-reason methods followed \citep{press2023selfask,trivedi2023ircot}; BrowseComp \citep{wei2025browsecomp} and its fixed-corpus variant BrowseComp-Plus \citep{chen2025browsecomp} push the setting to questions whose search terms do not appear in the question, so answering requires discovering an entity the question never names and rejecting documents that look relevant but are not. Deep-research agents \citep{jin2025searchr1,li2025websailor,tongyi2025deepresearch} handle this with long tool-calling trajectories on large models, which is slow and expensive, and are evaluated on the final answer, which conflates retrieval with generation.

T-Search targets the retrieval stage in isolation. Given a question and a search backend, it runs a bounded multi-round search and returns a ranked list of evidence chunks, each with a one-line reason, for a downstream model of the user's choice to answer from. It never writes the answer, so one retriever serves many generators; it reaches documents only through the search tool, so the backend can be swapped without retraining; and it returns chunk identifiers, so its quality is measured directly as recall of gold evidence rather than through an answer judge. Three design choices make this work: short, bounded \emph{rounds} in a ReAct loop \citep{yao2023react} that carry forward only explicitly saved evidence, so tool output does not accumulate and degrade the model's judgment \citep{hong2025contextrot}; search difficulty engineered into the training data and verified by adversarial solvers; and a recall reward on chunk identifiers with no judge model, optimized with GSPO \citep{zheng2025gspo}.

Our contributions are:\footnote{Model, benchmarks, and harness: \url{https://huggingface.co/collections/t-tech/t-search}. Live demo: \url{https://t-search.app/}. Demo video: \url{https://youtu.be/_nkS_VJraVA?si=2LsKVReFlDfpYjtT}.}
\begin{itemize}\setlength{\itemsep}{2pt}
  \item T-Search, an open-weight (Apache-2.0) agentic retriever with 35B total and ${\sim}$3B active parameters that outperforms larger open models on average over seven fixed-index benchmarks and lies on the latency--quality frontier (Section~\ref{sec:evaluation});
  \item TRuST, a hand-built native-Russian hard-search benchmark, and SynthComp-En/Ru, both with fixed indices and gold evidence annotations (Appendix~\ref{app:benchmarks});
  \item A web playground that runs T-Search over TRuST or BrowseComp-Plus using dense or BM25 \citep{robertson2009probabilistic} retrieval, or over the live web, replays the search as a timeline of queries, saved chunks, and rejected actions with the rule each broke, and scores benchmark questions with the Recall@10 of Section~\ref{sec:evaluation} (Section~\ref{sec:demo}).
\end{itemize}

The model, with FP8 and NVFP4 quantizations, and the harness\footnote{\url{https://github.com/turbo-llm/t-search-harness}} are released under Apache-2.0 and the benchmarks under ODC-BY.

\section{Related Work}
\label{sec:related}

\paragraph{Search agents.}
Search-R1 \citep{jin2025searchr1}, WebDancer, WebSailor, WebShaper \citep{wu2025webdancer,li2025websailor,tao2025webshaper}, and Tongyi DeepResearch \citep{tongyi2025deepresearch} are trained end-to-end over a growing transcript and judged on the final answer, which conflates retrieval with generation. The concurrent Harness-1 \citep{jiang2026harness1reinforcementlearningsearch} is closest to T-Search: a 20B retrieval subagent trained with RL inside a harness that returns curated evidence rather than an answer. The two divide labor between environment and model in opposite directions. Harness-1 keeps a candidate pool, evidence graph, and verification records in the environment and re-renders them into the context every turn, shaping behavior through a multi-term reward; T-Search carries saved evidence, round summaries, coverage declarations, and goals into each new round, so a chunk survives only if the policy saves it, and a single trajectory-level recall reward suffices. Harness-1 is also English-only and its released search tool relies on an external embedding API, whereas T-Search is bilingual and treats the backend as a swappable component. Section~\ref{sec:evaluation} evaluates all six baseline agents over the same fixed indices.

\paragraph{Benchmarks and context.}
We adopt the fixed-index, gold-evidence design of BrowseComp-Plus \citep{chen2025browsecomp} and SealQA \citep{pham2025sealqa} and contribute TRuST, the first native-Russian hard-search benchmark, and the bilingual SynthComp. Open Russian-language releases have so far targeted general reasoning and efficient inference \citep{stoianov2026tpro} rather than multi-step retrieval. Short clean rounds with model-controlled saving sit between external memory \citep{packer2023memgpt} and summarization, avoiding both re-fetching and lossy rewrites where added context degrades judgment \citep{hong2025contextrot}.

\section{System Overview}
\label{sec:system}

\begin{figure*}[!t]
\centering
\includegraphics[width=0.9\linewidth]{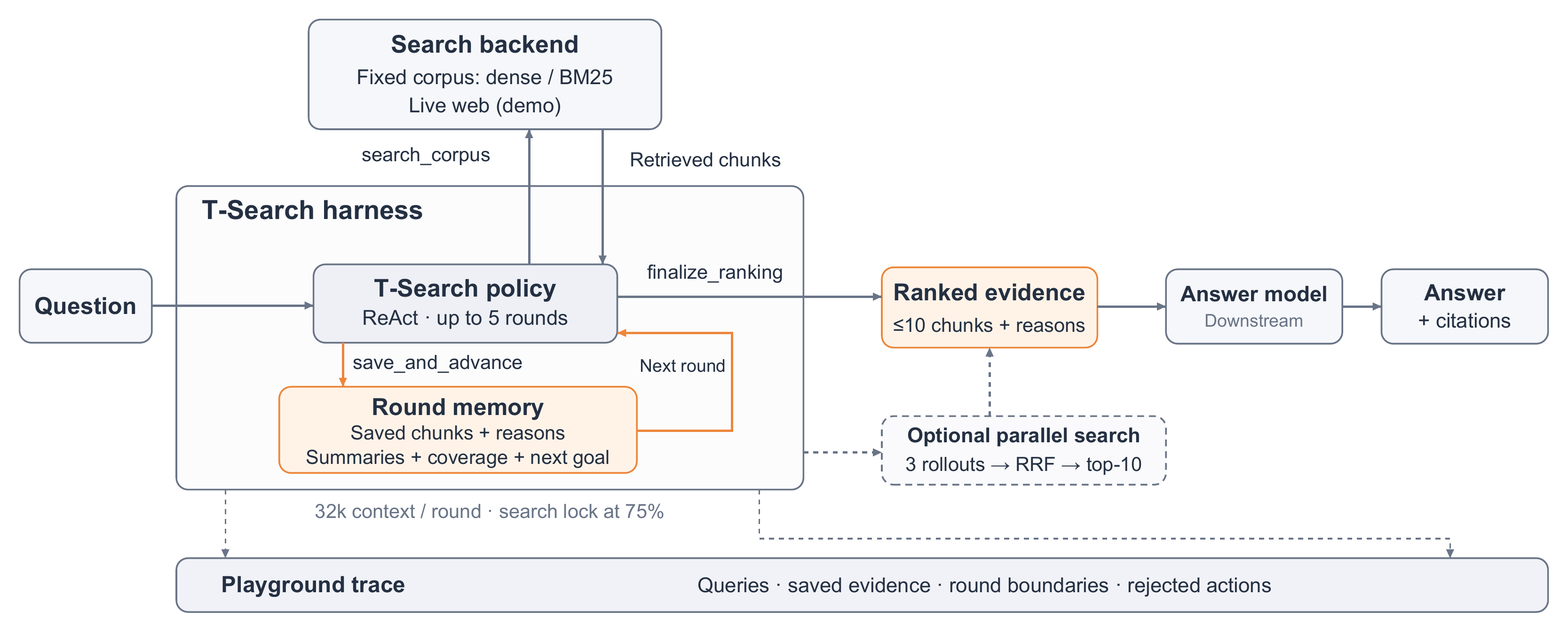}
\caption{T-Search and its playground. The policy searches a selected backend, carries explicit memory between rounds, and returns ranked evidence. Optional parallel rollouts are fused with RRF. The playground displays the search trace, while a separate downstream model generates the answer. Fixed-index evaluation uses corpus backends; live web search is available in the demo.}
\label{fig:rounds}
\end{figure*}

\paragraph{Search and memory.}
Given a question and a search backend, the policy operates through three tools: \texttt{search\_corpus}, \texttt{save\_and\_advance}, and \texttt{finalize\_ranking}. Within a round, it issues queries, inspects returned chunks, and follows intermediate entities or unresolved constraints. To continue in a fresh context, it selects chunks to retain with a reason for each, records which parts of the question are covered or unresolved, and supplies a round summary and a goal for the next round. The harness constructs the next input from the original question, saved chunks, round summaries, coverage state, and next-round goal (Figure~\ref{fig:rounds}). The preceding tool transcript and full reasoning history are discarded.

\paragraph{Budgets and stopping.}
The default configuration permits up to five rounds, each with a context budget of about 32k tokens. At 75\% of that budget, search is locked. Advancing requires at least five searches and one saved chunk. Before the last allowed round, finalization is rejected if at least half of the declared question parts remain unresolved. In the last round, advancing is disabled and the coverage restriction on finalization is lifted. Appendix~\ref{app:harness} gives the complete tool schemas, prompts, and validation rules.

\paragraph{Output.}
On successful finalization, the policy supplies a ranking with reasons and coverage declarations, using saved evidence or chunks retrieved in the current round. The harness exports at most ten chunks; Appendix~\ref{app:harness} describes fallback behavior when no finalization is accepted. Three parallel rollouts can be combined by reciprocal rank fusion \citep{cormack2009rrf}, followed by truncation to ten chunks; Section~\ref{sec:scaling} evaluates the latency--recall trade-off.

\section{Demonstration}
\label{sec:demo}

\begin{figure}[t]
  \centering
  \includegraphics[width=0.9\columnwidth]{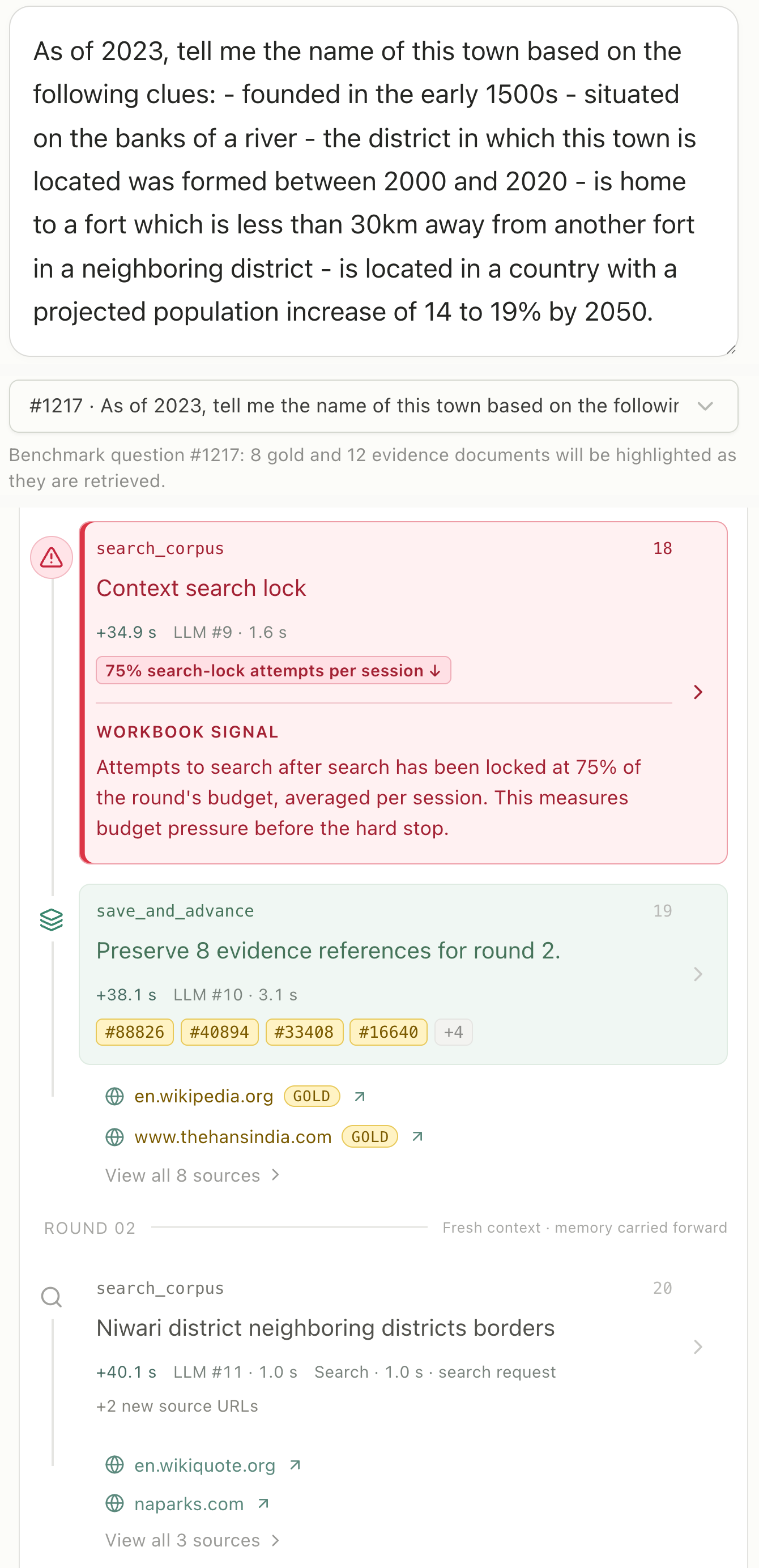}
\caption{A run in the playground on BrowseComp-Plus question \#1217 over the Qwen3-Embedding-8B \citep{zhang2025qwen3} index. Top: the question and the benchmark picker, with gold and evidence documents marked for badging. Below, the trace around a round boundary: a search after the 75\% lock is rejected and flagged with the rule it broke; the save closes round one with eight chunks, the sources shown badged as gold documents; round two opens with a fresh context and its first query with the sources it returned.}
  \label{fig:playground}
\end{figure}

The playground at \url{https://t-search.app/} runs the released checkpoint inside the harness of Section~\ref{sec:system} and shows what the harness sees: every query and its results, every chunk the policy kept and why, the coverage state at each round boundary, and every action the guardrails rejected (Figure~\ref{fig:playground}).

\paragraph{Architecture.}
The harness runs on the server side of a Cloudflare Worker application: it holds the released prompt and tool schemas, enforces the rules of Section~\ref{sec:system}, calls the model over an OpenAI-compatible API with the official sampling settings, and journals progress so that a run survives a closed tab and can be stopped. The FP8 checkpoint is served by SGLang \citep{zheng2024sglang} on one serverless H200; GPU and index containers start on demand in under a minute and stop after five idle minutes. The backends are a live web-search API (five snippets per query) and a CPU service over TRuST (324 Russian questions, 2.7M passages) and BrowseComp-Plus (830 English questions, 100k pages), each with the benchmark's own Qwen3-Embedding-8B index and with BM25. A separate instruct model (Qwen3.6-35B-A3B; \citealp{qwen2026qwen36_35}, reasoning off) writes the answer last from the ranked snippets through a forced \texttt{submit\_answer} tool, with citations verified against observed sources.

\paragraph{Interface.}
The question box has selectors for the index and a time budget; a corpus index also lists its benchmark questions. The trajectory is a timeline of tool actions split by round dividers: queries with their new chunks and sources, saves with the kept chunks, reasons, and coverage lists, and the final ranking. Rejected actions stay in place, flagged with the rule they broke, and every action opens to its raw arguments and result. Three panels derive from the visible events: \emph{context growth} (estimated prompt tokens per call against the 32k budget, dropping at each round reset), \emph{memory contribution} (ranked sources that survived a reset only because the policy saved them), and \emph{findings chain} (query wording that earlier results supplied, i.e.,\ the bridge entities). For a benchmark question, gold and evidence passages are badged as they arrive and the outcome line reports how many were seen and how many were ranked, the metric of Section~\ref{sec:evaluation} for one question; the reference answer can be revealed afterwards, and the model sees neither the badges nor the answer.

\paragraph{Walkthrough.}
The screencast replays a completed T-Search run on a BrowseComp-Plus question using the dense index. The policy searches from several angles and gold badges appear as the right passages surface; a save closes the round and the context plot drops back to the saved chunks; finalization returns the ranking, which the outcome line scores against the reference evidence documents, and the answer model writes a cited answer to compare with the reference. The screencast concludes by showing the available search backends and switching to live web search.

\section{Training Recipe}
\label{sec:training}

We fine-tune Qwen3.6-35B-A3B on synthetic search tasks in two stages: supervised fine-tuning (SFT) on teacher trajectories and reinforcement learning, both inside the harness of Section~\ref{sec:system}. Tasks are generated over local corpora, so the policy is trained in the loop it is evaluated in.

\paragraph{Task synthesis and validation.}
Each task contains a question, a short reference answer, annotated support chunks, and a fixed index with distractors. Questions are generated natively in each language rather than translated between them, the same choice \citet{tsymboi2026production} make for Russian function-calling data after finding that native generation matters. A taxonomy specifies the search mechanic, question form, answer type, and distractor properties. Compatible combinations are sampled as generation instructions, with weights adjusted to balance coverage across search behaviors. To support generation with a small local model, we preselect connected chunk subgraphs using named-entity annotations, mention frequencies, and co-occurrence statistics, emphasizing niche entities and less obvious links. Facts are extracted and paraphrased into an intermediate representation; the composer does not see the original quotations, reducing direct lexical clues. It extends a question one hop, entity, or constraint at a time, checking answerability and whether the added support introduces a necessary search step.

An adversarial loop then probes shortcuts beyond these local checks. It tests answerability from the full support set, answer or intermediate-entity leakage, closed-book solvability, and direct retrieval of the required evidence using the whole question as a query. Each support chunk is removed in turn to test its necessity, and superfluous chunks are pruned. Finally, the composer model runs as a search agent; answer correctness and evidence recall are used to identify tasks it solves too easily. Checks proceed from cheaper probes to full agent runs. Feedback names the redundant chunk, leaked clue, or successful shortcut query, and the composer revises and resubmits the task. About 40k of 67k candidates survive this process (Appendix~\ref{app:factory}).

\paragraph{Data split and teacher trajectories.}
Each language contributes about 8k SFT questions and a separate pool of 2k RL questions. GLM-5.1 \citep{zai2026glm51} collects teacher trajectories. Each is split into self-contained rounds: the input contains the question and the harness state at the start of the round; the target contains the round's search actions and continuation or finalization decision.

\paragraph{Round selection and mixture.}
Invalid tool actions are masked from the loss while useful recovery after tool feedback is retained. A save round must add previously unsaved support; finalization rounds are selected by evidence coverage. Productive earlier rounds can therefore be retained when a trajectory ends with poor coverage, preserving useful supervision on difficult questions. Selection balances round number, ending type, task difficulty, task mechanic, and evidence completeness. Frequent cells are capped, while rare useful cells are retained, yielding about 11k rounds per language. A quarter of the mixture targets retriever robustness with semantic, BM25, and hybrid search, varying embedding backends and retrieval depth (Appendix~\ref{app:sft}).

\begin{table*}[t]
\centering\small
\setlength{\tabcolsep}{4.5pt}
\begin{tabular}{@{}llrrrrrrrr@{}}
\toprule
& & \multicolumn{2}{c}{BrowseComp-Plus} & \multicolumn{2}{c}{SealQA} & \multicolumn{2}{c}{SynthComp} & & \\
\cmidrule(lr){3-4}\cmidrule(lr){5-6}\cmidrule(lr){7-8}
Model & Harness & En & Ru & En & Ru & En & Ru & TRuST & Avg. \\
\midrule
\textbf{T-Search} ($N{=}3$) & T-Search & \textbf{72.65} & \textbf{62.93} & \textbf{66.08} & \textbf{61.98} & \textbf{58.52} & \textbf{58.00} & \textbf{49.12} & \textbf{61.33} \\
\textbf{T-Search} ($N{=}1$) & T-Search & 65.35 & 55.95 & 61.16 & 57.72 & 54.52 & 53.13 & 43.92 & 55.96 \\
GLM-5.1 & T-Search & 64.32 & 58.18 & 55.49 & 53.21 & 51.69 & 51.71 & 43.11 & 53.96 \\
GLM-5.2$^{1}$ & T-Search & 63.01 & 52.54 & 55.30 & 54.69 & 52.29 & 49.37 & 37.07 & 52.04 \\
Kimi-K2.6$^{2}$ & T-Search & 60.71 & 49.76 & 56.86 & 52.46 & 48.25 & 47.06 & 42.39 & 51.07 \\
Qwen3.6-27B$^{3}$ & T-Search & 56.69 & 46.81 & 54.01 & 48.95 & 46.82 & 46.99 & 38.12 & 48.34 \\
DeepSeek-V4-Flash$^{4}$ & T-Search & 53.27 & 46.30 & 55.70 & 51.12 & 44.83 & 45.68 & 40.40 & 48.19 \\
Qwen3.5-397B-A17B$^{5}$ & T-Search & 53.48 & 44.06 & 51.68 & 48.33 & 47.38 & 46.91 & 38.57 & 47.20 \\
Qwen3.5-122B-A10B$^{6}$ & T-Search & 47.81 & 40.15 & 51.57 & 44.74 & 42.11 & 42.25 & 30.55 & 42.74 \\
Qwen3.6-35B-A3B (base) & T-Search & 43.66 & 38.58 & 46.07 & 43.26 & 41.82 & 43.88 & 33.53 & 41.54 \\
gemma-4-26B-A4B-it$^{7}$ & T-Search & 35.32 & 27.51 & 45.13 & 39.05 & 36.24 & 34.77 & 21.75 & 34.25 \\
\midrule
Harness-1 & Harness-1 & 40.42 & 31.34 & 42.92 & 40.57 & 39.26 & 36.32 & 18.96 & 35.68 \\
\midrule
Tongyi DeepResearch & Tongyi & 58.48 & 38.86 & 70.84 & 57.30 & 60.76 & 43.90 & 37.71 & 52.55 \\
WebSailor-32B & WebSailor & 30.50 & 19.70 & 49.83 & 47.00 & 48.17 & 39.32 & 23.99 & 36.93 \\
WebDancer-32B & WebDancer & 23.65 & 16.30 & 52.12 & 49.98 & 37.26 & 34.14 & 18.90 & 33.19 \\
WebShaper-32B & WebDancer$^*$ & 23.66 & 17.76 & 48.53 & 48.85 & 37.76 & 35.52 & 18.69 & 32.97 \\
Search-R1-32B & Search-R1 & 9.32 & 6.75 & 22.09 & 20.52 & 19.25 & 13.30 & 7.47 & 14.10 \\
\bottomrule
\end{tabular}
\par\smallskip
\begin{minipage}{\textwidth}
\footnotesize
Model sources:
$^{1}$\citealp{zai2026glm52};
$^{2}$\citealp{moonshot2026kimi};
$^{3}$\citealp{qwen2026qwen36_27};
$^{4}$\citealp{deepseek2026v4flash};
$^{5}$\citealp{qwen2026qwen35_397};
$^{6}$\citealp{qwen2026qwen35_122};
$^{7}$\citealp{gemma2026gemma4}.
\end{minipage}
\caption{Retrieval results (\%).
Top and middle blocks report final-set Recall@10;
the bottom block reports trajectory recall.
Bold marks the highest final-set Recall@10 in each column.
$N{=}3$ denotes three rollouts fused with RRF.
$^*$WebShaper uses the WebDancer inference recipe.}
\label{tab:main}
\end{table*}
\begin{figure*}[t]
\centering
\includegraphics[width=0.75\linewidth, trim=0 0 62 8, clip]{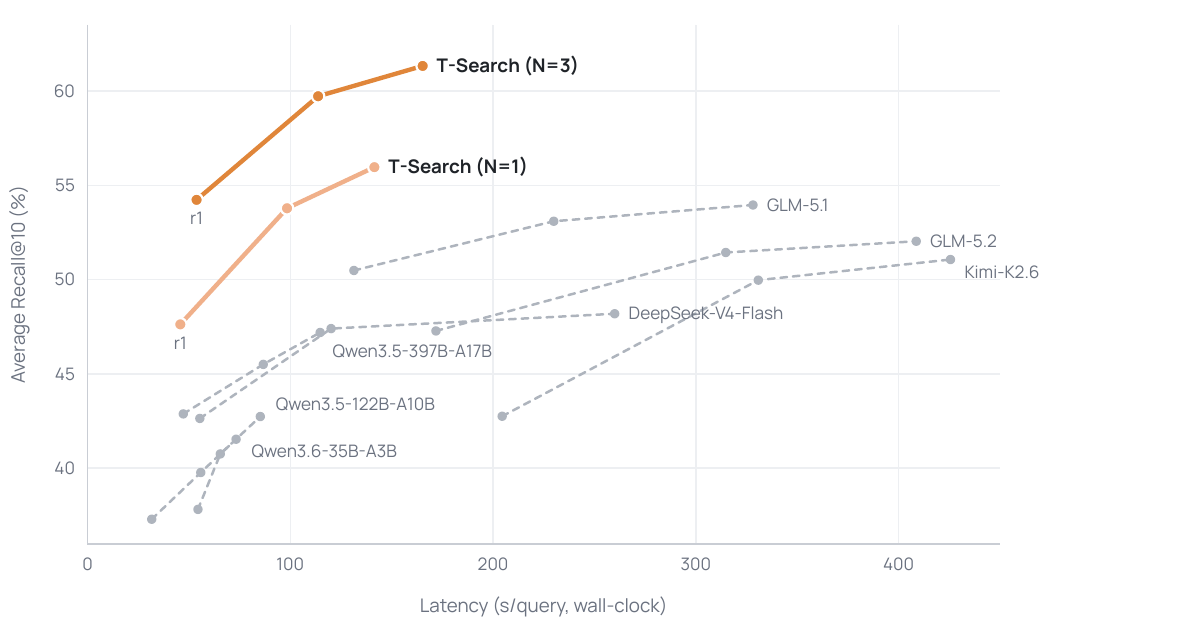}
\caption{Test-time scaling. Each curve sweeps the round cap $r \in \{1,3,5\}$ for one model in the T-Search harness, left to right; the T-Search $N{=}3$ curve additionally fuses three parallel rollouts with RRF. Latency is per-query wall-clock time of the complete agent loop on 35 questions; recall is the mean final-set Recall@10 across benchmarks.}
\label{fig:scaling}
\end{figure*}

\paragraph{Reinforcement learning.}
The RL stage starts from the merged SFT checkpoint and trains on the separate RL question pool for each language. Each trajectory receives a single scalar reward, the recall of the reference support set in the finalized ranking, computed from chunk identifiers without a judge model (Eq.~\ref{eq:reward}); there is no separate reward for saving and no per-round shaping. Recall is preferred over precision-based objectives, which in preliminary runs led to premature finalization with incomplete evidence. We optimize with GSPO \citep{zheng2025gspo}; training dynamics and hyperparameters are reported in Appendix~\ref{app:rl}.

\paragraph{Merging.}
Each stage is run separately for English and Russian, and the branches are merged back before the next one: DARE \citep{yu2023dare} combines the two SFT branches into the checkpoint that initializes both RL branches, and SLERP \citep{shoemake1985slerp,goddard2024mergekit} combines the two RL branches into the released model (Figure~\ref{fig:recipe}, Appendix~\ref{app:rl}); the same two-stage structure is used for per-capability experts in \citet{tsymboi2026production}. The release is thus a single bilingual model.

\section{Evaluation}
\label{sec:evaluation}
\label{sec:benchmarks}

\paragraph{Setup.}
We evaluate on BrowseComp-Plus, the Seal-Hard subset of SealQA, our Russian translations of both, and our newly constructed TRuST and SynthComp benchmarks. TRuST contains 324 manually authored Russian questions, while SynthComp contains 395 synthetic questions each in English and Russian. Construction, localization, and scoring are detailed in Appendix~\ref{app:benchmarks}, including SynthComp's separation from training data in Appendix~\ref{app:synthcomp}.

For the main comparison, all models search the same fixed corpora with Qwen3-Embedding-8B, without reranking and with search snippets capped at 512 tokens. T-Search and general-purpose baselines, including its base model, use the T-Search harness with a five-round cap. Harness-1 and generative search agents use adapted released harnesses.

The top and middle blocks of Table~\ref{tab:main} report final-set Recall@10. The bottom block reports trajectory recall over all unique search results retrieved during a rollout, which is not directly comparable to final-set Recall@10 (Appendix~\ref{app:search-agent-eval}). Recall uses reference documents for BrowseComp-Plus and TRuST, annotated evidence chunks for our SealQA adaptation, and all reference evidence chunks for SynthComp. Each score column averages recall over the questions in the corresponding benchmark and language. Avg.\ is the arithmetic mean of the preceding score columns in Table~\ref{tab:main}.

\paragraph{Results.}
Table~\ref{tab:main} presents the main results. At a single rollout, T-Search improves over its base model on every benchmark and achieves the highest average Recall@10 among single-rollout configurations in the T-Search harness (55.96). It outperforms all general-purpose baselines except on Russian BrowseComp-Plus, where GLM-5.1 leads. Fusing three rollouts increases recall on every benchmark and raises the average to 61.33. With fusion, T-Search achieves the highest final-set Recall@10 on all benchmarks. Among the generative agents, Tongyi DeepResearch has the highest average trajectory recall. Recall increases from the base model to SFT and from SFT to the released model on every benchmark (Appendix~\ref{app:diagnostics}, Table~\ref{tab:training_stages}).

\paragraph{Test-time scaling.}
\label{sec:scaling}
The round cap and the number of fused rollouts control the latency--recall trade-off. Figure~\ref{fig:scaling} varies the round cap over $\{1,3,5\}$ in the T-Search harness. Latency is measured on 35 questions using a protocol inspired by \citet{chang2026karl}, while recall uses all evaluation questions (Appendix~\ref{app:diagnostics}). In one round, T-Search achieves an average recall similar to Qwen3.5-397B-A17B in five rounds, at 40\% of its latency. At three rounds, it approaches GLM-5.1 at five rounds using less than a third of its latency. Parallel rollout fusion further improves recall at the cost of additional inference work.

\paragraph{Retriever robustness.}
\label{sec:robustness}
T-Search supports all tested retrievers with the same checkpoint and agent inference settings ($N{=}1$), including alternative dense retrievers, BM25, and LLM reranking. Average Recall@10 ranges from 49.38 with Qwen3-Embedding-0.6B to 62.87 with an LLM reranker on top of the default retriever, compared with 55.96 for the default retriever (Table~\ref{tab:backends}, Appendix~\ref{app:diagnostics}). Reranking improves recall on every benchmark. Retriever choice still matters: BM25 performs best on SynthComp and TRuST, but worst on BrowseComp-Plus among the tested retrievers.

\section{Conclusion}
\label{sec:conclusion}

T-Search treats evidence gathering as a component in its own right. A bounded multi-round harness keeps the working window short while the search runs long, a synthetic data factory builds search difficulty into the corpus and verifies that difficulty by trying to shortcut it, and round-sliced SFT followed by GSPO on a recall reward turns a 35B mixture-of-experts into a retriever that averages 56.0 Recall@10 over seven English and Russian benchmarks, 14.4 points above its base under the same five-round cap and ahead of considerably larger open models; three-rollout fusion raises this to 61.3. Because the agent reaches documents only through a tool, practitioners can swap the backend, point the harness at their own index, and keep the generator they already run.

\section*{Limitations}
\paragraph{The harness is part of the model.}
The rounds, the three-tool contract, and the guardrails are what the policy learned to operate inside, so quality can shift in a different loop. We recommend running the released harness and validating on your own index before relying on the reported numbers.

\paragraph{Evidence, not answers.}
T-Search returns a ranked evidence set with justifications and leaves generation to a downstream model, which is the intended design but means it cannot be deployed or evaluated as a standalone question-answering system. On general-purpose benchmarks, we did not observe noticeable differences from the base model.

\section*{Ethical Statement}

\paragraph{Data provenance.}
TRuST questions were written by 15 annotators employed by T-Tech and are grounded in publicly accessible Russian-language web pages. The released index contains 512-token chunks of those pages together with their source URLs, is distributed under ODC-BY for research use, and we honor removal requests from page owners. Every person named in a question or its evidence appears in a public role; we verified this for the whole benchmark after collection. SynthComp-En is built from Essential-Web \citep{essentialai2025web}, while SynthComp-Ru uses Russian-language sources including FineWeb2 \citep{penedo2025fineweb2} and Russian Wikipedia. The benchmark releases are licensed under ODC-BY; source texts remain subject to their original terms.

\paragraph{Model-generated data and text.}
Training tasks, teacher trajectories, and the difficulty checks for TRuST were produced by LLMs (the composer model, GLM-5.1, and GPT-5.4); every task is validated for answerability against its annotated support chunks, which come from the corpus rather than from a model. The per-chunk reasons that T-Search returns are model-generated and can be wrong; they help a reader locate evidence and do not replace reading it.

\paragraph{Intended use and misuse.}
T-Search returns evidence rather than answers and is optimized for recall, so a returned list can include chunks that turn out not to support the answer; downstream systems must verify claims against the returned text. Like any web search agent, it could be used to aggregate information about individuals. In the public playground the live-web retriever passes the five snippets returned by the search API per query directly to the agent and never fetches page content, so it exposes no information that the underlying search engine does not already return for the same queries; the API's content policies apply, runs are bounded by the round cap and a time budget, no conversations are stored, and run journals expire after two hours. The released harness lets operators restrict the backend to their own corpus, which is the intended deployment.

\paragraph{Licensing, coverage, and compute.}
The base model is released under Apache-2.0, and the harness and the released checkpoints, including the FP8 and NVFP4 quantizations, are distributed under the same license; benchmarks are ODC-BY. The model is trained and evaluated only in English and Russian; behavior in other languages is untested. SFT and RL for both languages used four nodes of eight H100 GPUs for about 3k GPU-hours in total; the demo serves one on-demand GPU that stops after five idle minutes.

\bibliography{custom}
\clearpage
\appendix
\section{Synthetic data and SFT details}
\label{app:factory}

\begin{figure*}[!tp]
\centering
\includegraphics[width=\textwidth]{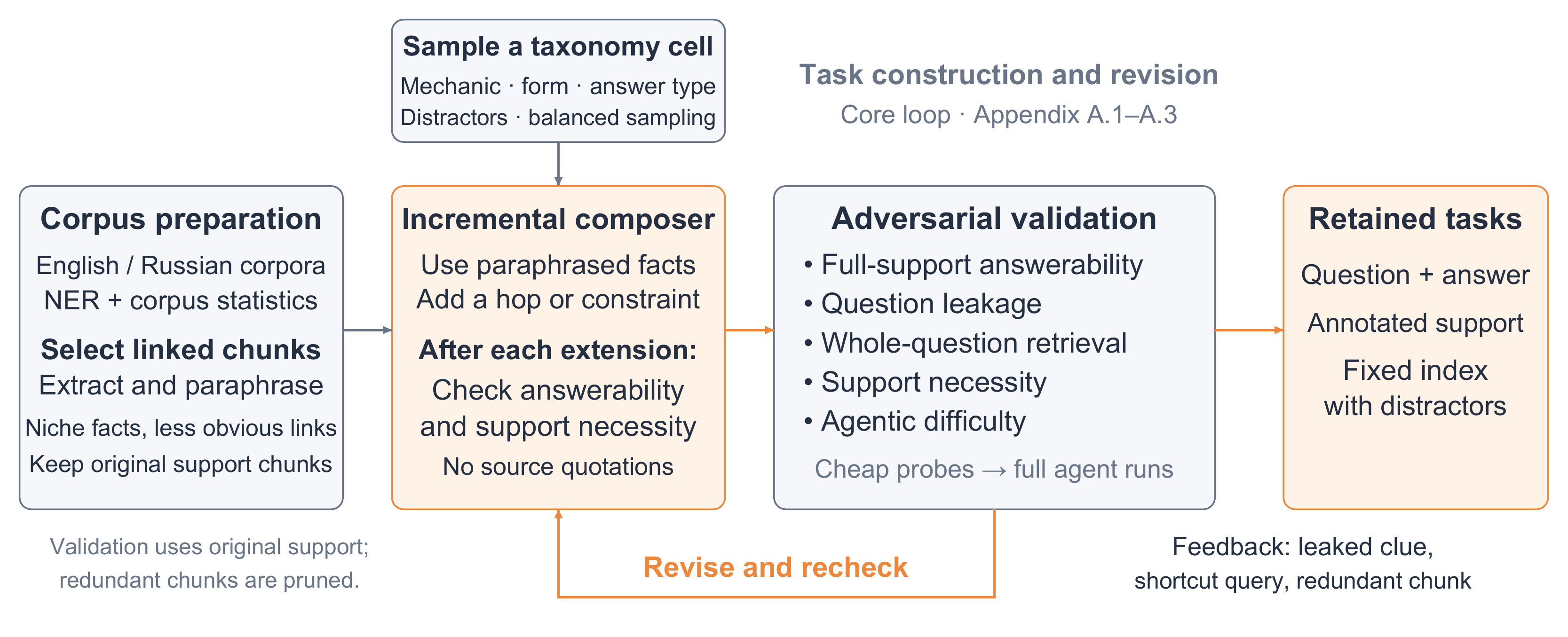}
\caption{Synthetic task construction and revision. A taxonomy specification and paraphrased corpus facts guide incremental composition. Original support chunks remain available for validation. The composer model also acts as the search agent in the difficulty check. Actionable feedback returns to the composer for revision and rechecking; retained tasks provide questions, reference answers, annotated support, and fixed retrieval indices.}
\label{fig:synthetic_pipeline}
\end{figure*}

\subsection{Task specification and controlled diversity}
Figure~\ref{fig:synthetic_pipeline} summarizes the core construction and revision loop.
The reference answer and support annotations serve different purposes: the answer is used to validate the question, while the support set enables evidence-based checks and trajectory selection. The trained retriever returns evidence rather than answer text.

Taxonomy cells describe intended search behavior, not merely subject matter. For example, linking intermediate entities differs from intersecting independent constraints or gathering candidates for comparison; temporal reasoning and calculations impose further requirements on the supporting facts. Question form, answer type, and distractor properties provide additional controls. Only compatible combinations are supplied to the composer, and sampling weights are adjusted to cover underrepresented combinations. A specification expresses the intended task, but does not establish that its solution actually requires the intended searches. That is tested after composition.

\subsection{Corpus-grounded, incremental composition}
\paragraph{Selecting the factual basis.}
The initial exploratory generator, inspired by Context-1 \citep{bashir2026context1}, searched the index for related facts but tended to produce entity-chain questions and choose facts accessible through broad queries or model knowledge. The final pipeline instead prepares candidate subgraphs before question construction. Connections may involve entities, events, dates, or other constraints. Niche facts and less obvious links help reduce trivial starting points, while preselection relieves the small composer of having to discover a suitable factual chain by exploring the entire index.

\paragraph{Controlling lexical shortcuts.}
The composer receives paraphrased facts rather than source quotations. A distinctive phrase copied into a question could itself serve as an effective retrieval query even when the wording appears to require multiple reasoning steps. Paraphrasing reduces this risk but may distort a fact, so correctness is checked against the source support chunks. The intermediate representation is an input to composition, not a replacement for the source evidence used in validation.

\paragraph{Local and task-level checks.}
After each extension, answerability and the necessity of the added support are checked before composition continues. These checks localize a broken chain or unnecessary constraint to an individual extension. Validation of the completed question then attempts to bypass the intended chain. A locally valid extension may still leave a shortcut through the question wording, a broad retrieval query, or facts already known to the model.

\subsection{Adversarial validation and revision}
The checks use different inputs and test different failure modes:
\begin{enumerate}
\item \textbf{Full-support answerability.} Given the question and all annotated support chunks, a solver must be able to derive the reference answer. This checks that composition and paraphrasing have not produced an unsupported task.
\item \textbf{Question leakage.} The wording is examined for direct or near-direct disclosure of the answer or a key intermediate entity. Revealing an intermediate search target can collapse a multi-step task even if the final answer is not stated.
\item \textbf{Whole-question retrieval.} The complete question is submitted as a search query and the top results are inspected for the required evidence. This tests whether the task can be bypassed by retrieving directly from its wording.
\item \textbf{Support necessity.} Each annotated support chunk is removed separately and answerability is checked again. A chunk that does not affect the tested solver's ability to answer is pruned instead of being counted as an additional reasoning step.
\item \textbf{Closed-book solvability.} The question is tested without retrieved evidence to identify answers available from model knowledge rather than the intended search process.
\item \textbf{Agentic difficulty.} The composer model attempts the question through the search harness. Answer correctness and evidence recall are examined together to identify tasks that remain too easy. A successful agent run is not by itself the acceptance criterion.
\end{enumerate}
Failures return actionable feedback to the composer, such as a support chunk that can be removed or a query that exposes the answer directly. Revised tasks are checked again. The checks are complementary: answerability with all supports does not establish their individual necessity, and failure of the whole-question query does not rule out another shortcut.

Support necessity is relative to the tested solver and context. Failure to answer after removing a chunk does not prove global minimality of the support set or the absence of every alternative solution. The aggregate retention count in Section~\ref{sec:training} measures the output of the overall process, not the contribution of any single check.

\subsection{Teacher trajectories and SFT selection}
\label{app:sft}
\paragraph{Round boundaries and supervision.}
For rounds after the first, the incoming state contains saved evidence with reasons, summaries of earlier rounds, coverage declarations, and the next-round goal (Appendix~\ref{app:agent_prompts}). The retained example therefore includes that state rather than reconstructing the entire preceding tool history. Round boundaries expose the student to initial, intermediate, and terminal search states, and allow selection at a finer granularity than whole trajectories.

\paragraph{Masking and quality checks.}
An invalid action is excluded from the supervised loss, but the resulting tool error can provide context for a useful correction. The continuation is not automatically discarded because an earlier action was invalid. Additional checks inspect tool-call structure and required arguments and remove looping or unjustifiably long reasoning. Local masking concerns which actions are supervised; round selection concerns whether the remaining search behavior is useful.

\paragraph{Progress-based continuation selection.}
Let $E$ be the task's reference support set, $I$ the chunk identifiers saved in the incoming state, and $S$ those saved at the end of a round. For a round ending in \texttt{save\_and\_advance}, the progress criterion is
\[
\left|(S\cap E)\setminus I\right|>0.
\]
Saving only previously preserved evidence, or issuing more queries without adding required support, does not satisfy this criterion.

\paragraph{Finalization and difficult trajectories.}
For a round ending in \texttt{finalize\_ranking}, selection considers coverage of the reference evidence in the returned set. A weak terminal round can be removed while useful earlier rounds from the same trajectory remain. Filtering only by a trajectory's final recall would also discard those productive rounds, biasing supervision toward questions that the teacher can already solve well. The continuation criterion above and finalization coverage are SFT selection statistics; they do not define RL rewards.

\paragraph{Mixture construction and retriever diversity.}
Initial rounds are common, while later rounds that still acquire useful evidence are rarer. Capping frequent mixture cells and retaining rare useful ones limits the dominance of easy initial states without discarding the latter. Balancing also accounts for task difficulty, search mechanic, and evidence completeness, so round position alone does not determine selection.

Retriever variation changes both which evidence is returned and the distribution of search-tool outputs seen during supervision. The collection settings vary the search method, embedding model, and retrieval depth. These data are used during SFT, whereas Appendix~\ref{app:diagnostics} evaluates retriever substitution with the final model held fixed. Section~\ref{sec:training} reports question counts, retained round counts, and the robustness-mixture proportion. Checks separating SynthComp from the SFT and RL pools are described in Appendix~\ref{app:synthcomp}.

\section{Reward and policy optimization}
\label{app:rl}

\begin{figure*}[!t]
\centering
\includegraphics[width=0.9\linewidth, trim=11 22 9 22, clip]{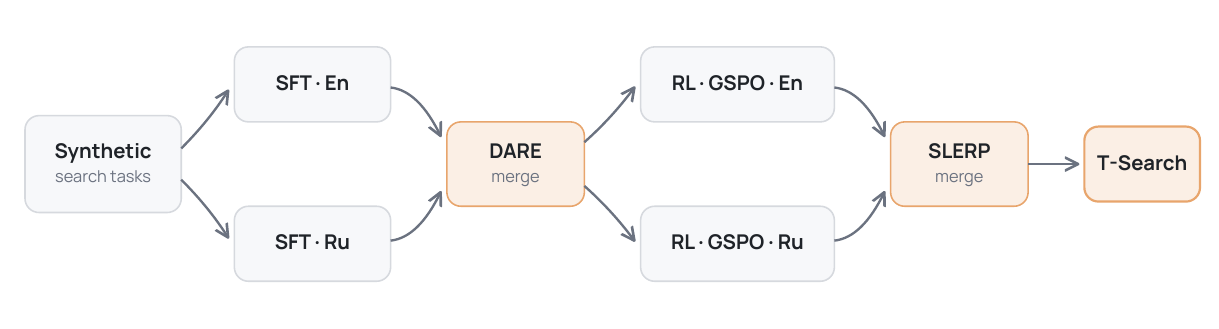}
\caption{The training recipe. English and Russian are trained separately at each stage: per-language SFT is combined with DARE \citep{yu2023dare}, per-language GSPO reinforcement learning with SLERP \citep{shoemake1985slerp,goddard2024mergekit}.}
\label{fig:recipe}
\end{figure*}
\paragraph{Trajectory-level learning.}
Starting from the merged SFT checkpoint, the policy runs its own complete searches: formulating queries, reading results, preserving chunks across rounds, and finalizing a ranked evidence set. RL therefore trains behavior under the states reached by the current policy, whereas SFT supervises selected teacher rounds. The reward compares the returned chunk identifiers with the task's reference chunk set $G$. A trajectory that finalizes the ranking $F$ receives
\begin{equation}
r = \frac{|F \cap G|}{|G|}.
\label{eq:reward}
\end{equation} This is the entire training signal: one scalar per trajectory, evaluated after finalization, with no per-save or per-round terms and no shaping. GSPO consumes it at the sequence level, as a single advantage per trajectory normalized within the group of rollouts sampled for the same question. The reward requires neither an LLM judge nor a comparison between generated answer text and a reference answer. It also does not use TRuST's document-level mapping: that mapping belongs to evaluation.

\paragraph{Why recall is the main signal.}
Exploratory precision/F1 objectives encouraged a conservative policy early in RL: search less, return a small set, and finalize with incomplete evidence. Recall instead rewards finding more of the required support. Precision and F1 remain diagnostics; the harness limits the number of returned chunks, so the output cannot grow without bound. Set-based recall gives no extra credit for returning a required identifier repeatedly. The reward measures evidence coverage, rather than correctness of a downstream generated answer or a preference for a particular ordering of already retrieved reference chunks.

\paragraph{Optimizer choice and language branches.}
GSPO uses sequence-level importance ratios rather than token-level ratios. This choice is motivated by discrepancies in expert routing between rollout generation and training for a mixture-of-experts policy. The adopted recipe does not record and replay routing decisions.

\begin{figure*}[!t]
\centering
\includegraphics[width=0.98\linewidth]{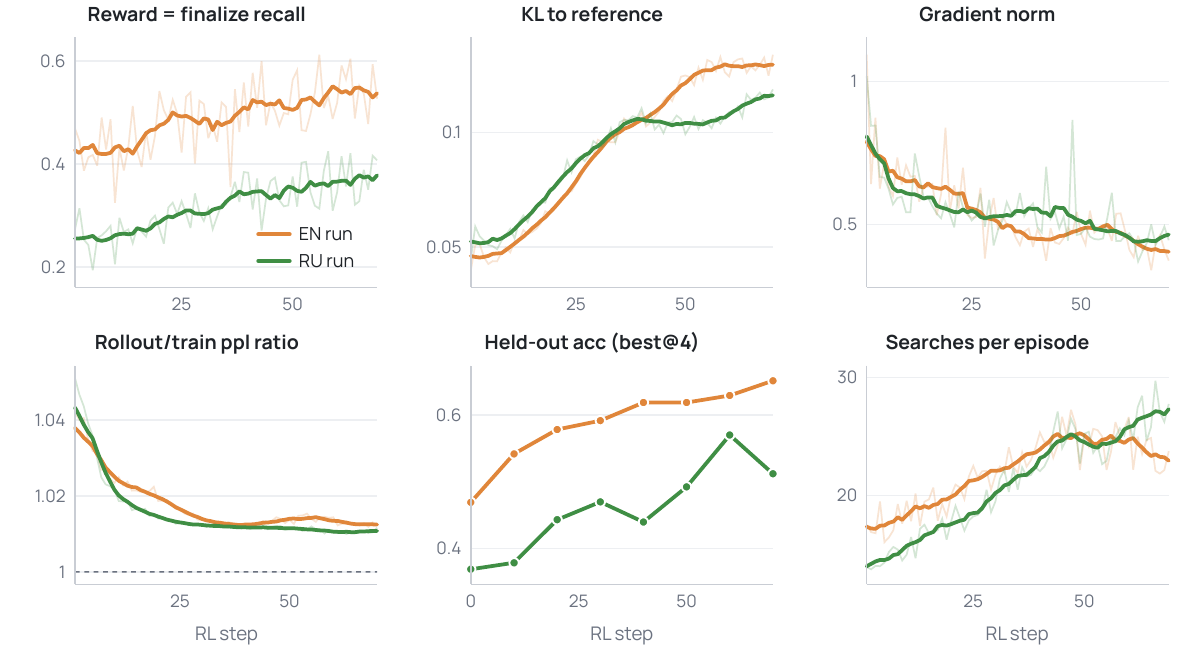}
\caption{RL training diagnostics for the two GSPO runs (English, Russian), trained separately before the SLERP merge. The reward panel plots the finalization recall of Eq.~\ref{eq:reward}, the only training signal; it rises over training, as does held-out accuracy; the number of search calls per episode grows; the gradient norm decays and the KL to the reference grows only modestly under the $10^{-3}$ penalty; and the rollout/train perplexity ratio converges toward 1, indicating that the vLLM rollout and Megatron training passes stay consistent without router replay. Held-out accuracy is best-of-4 answer accuracy on training-distribution questions held out from both SFT and RL, evaluated in the fixed round-based configuration; it is not a split of any evaluation benchmark.}
\label{fig:rl-diagnostics}
\end{figure*}

\paragraph{Training dynamics.}
Figure~\ref{fig:rl-diagnostics} tracks both runs over about 70 optimization steps. The mean finalization recall (Eq.~\ref{eq:reward}) rises from 0.44 to 0.53 in English and from 0.27 to 0.40 in Russian; the number of search calls per episode increases in both languages; best-of-4 held-out answer accuracy improves from 0.47 to 0.65 in English and from 0.37 to 0.51 in Russian; and the rollout-to-training perplexity ratio converges to about 1.01. Training runs on four nodes of eight H100 GPUs with verl \citep{sheng2025hybridflow} and a Megatron-Core backend at a learning rate of $10^{-6}$, a KL penalty of $10^{-3}$ to the reference policy, asymmetric clipping $(\varepsilon_{\mathrm{lo}},\varepsilon_{\mathrm{hi}})=(0.2,0.28)$, eight rollouts per question at temperature 0.7, and up to 28,672 response tokens per episode.

English and Russian SFT branches are first combined with DARE. That shared checkpoint initializes two language-specific RL branches, which are subsequently combined with SLERP. The comparison between the base and released models in Table~\ref{tab:main} measures the combined result of the full training recipe. Table~\ref{tab:training_stages} in Appendix~\ref{app:diagnostics} also reports the model after SFT. Because both trained models include language-branch merges, these results do not measure the effect of SFT or RL separately from merging, or compare alternative optimizers or merge operators.

\section{Benchmark construction and scoring}
\label{app:benchmarks}
We release TRuST (T-Tech Russian Search Test) and the English and Russian versions of SynthComp under ODC-BY, with fixed retrieval indices, evidence annotations, and hard negatives.
\subsection{Existing benchmarks and Russian localization}
Evaluation uses fixed indices rather than live web search, keeping the corpus available to each model fixed during evaluation. BrowseComp-Plus supplies the underlying evaluation methodology. SealQA's Seal-Hard questions are adapted to a fixed corpus: documents are split into 512-token chunks, which are annotated for whether they provide evidence needed for the answer. Thirty-three questions are removed because their documents lack useful evidence or contradict the reference answer, leaving 221 questions in each of the English and Russian versions.

Russian localization has two parts: translating the document corpus and translating question--answer pairs. Translation rules preserve named entities, numbers and units, document structure, formatting, and style. Answering from translated reference texts checks that the evidence still supports the reference answer. Such a check tests preservation of the evidence--answer relation; it does not establish equivalence of retrieval difficulty after translation. Translated tasks also remain distinct from tasks originally constructed from Russian sources, motivating the native Russian collections below.

\subsection{TRuST construction}

\paragraph{Design and coverage.}
TRuST contains 324 manually constructed Russian questions with short, unambiguous answers, created by 15 annotators from Russian-language web sources. Each example follows a hard-to-retrieve, easy-to-verify design: the answer should not be reliably recoverable from model knowledge or a single surface-level search, but should be unambiguously supported once the required evidence is found. The collection covers 8 topics: business and technology, sports, news and politics, science and academic publications, media and arts, regulation and law, geography, and history and archives. Five reasoning categories characterize the search difficulty: connecting multiple sources, using structured evidence, tracking changes over time, disambiguating entities, and comparing candidates. These categories target common challenges of the Russian web, including fragmented evidence, inconsistent entity naming, regional and archived pages, official and legal documents, and information distributed across multiple sites.

\paragraph{Question construction.}
Question construction starts from a seed fact or entity. Annotators progressively add properties that narrow the candidate set without directly revealing the answer, and then formulate a natural-language question with a short, definite answer. For each example, they record a final source confirming the answer together with the intermediate sources needed to connect the clues in the question to that answer, preferably across different websites. Questions themselves are authored manually, and LLMs are used only during validation rather than for question generation.

\paragraph{Difficulty and evidence validation.}
Each candidate passes several shortcut and difficulty checks. Annotators search the full question, paraphrased variants, and English reformulations to reject examples whose answer is surfaced directly by simple retrieval. GPT-5.4 independently attempts each question in closed-book, single-query, and full agentic-search settings. Questions solved from model knowledge or with a single-query search are returned for revision. For the agentic setting, we additionally use a 3-minute search-time threshold as a heuristic for selecting difficult examples.

The final 324 questions represent roughly 30\% of initially authored candidates, corresponding to approximately 1.1k candidate questions. About 30\% are filtered out by the difficulty checks, while a further 40\% fail subsequent evidence validation, primarily because the collected sources do not establish all required intermediate connections or provide insufficient evidence to recover the reference answer. Items are retained only when the question, reference answer, and collected evidence form a complete and consistent evidence chain.

\paragraph{Corpus construction.}
For indexing, each collected reference document is split into 512-token chunks, and the subset of chunks containing evidence needed to establish the answer is annotated explicitly. Each question is associated with an average of 6.01 gold documents and 6.32 annotated gold chunks, for 2,047 annotated gold chunks in total.

Finally, following BrowseComp-Plus, each question is decomposed into subqueries and search-engine results are used to mine hard negatives. These documents are plausible retrieval candidates but do not provide the evidence required to establish the answer, allowing the benchmark to test whether an agent can distinguish genuinely supporting evidence from realistic distractors.

\subsection{SynthComp in a shared index}
\label{app:synthcomp}
SynthComp-En uses Essential-Web \citep{essentialai2025web}; SynthComp-Ru draws on Russian web sources including FineWeb2 \citep{penedo2025fineweb2} and Russian Wikipedia. Candidate tasks generated with the training factory are pooled into a shared index for each language and undergo the additional validation described below.

\paragraph{Cross-task interference.}
A task validated in its original retrieval environment may acquire a shortcut when supports from other questions enter the index. The shared-index check therefore probes the neighborhood of both the original question and the queries from agent trajectories using the evaluation retriever. Candidates are excluded when those probes expose another task's supports or alternative documents that bypass the annotated evidence chain. The objective is to reduce cases where low annotated-evidence recall would penalize an agent for finding a different valid route. These probes cover the explored queries and retrieved neighborhoods, not every possible query or alternative proof.

\paragraph{Distractors and difficulty.}
Isolation is an admissibility check rather than a measure of difficulty. An isolated document can still be easy to find from a distinctive clue. Hard negatives are therefore mined again with the same subquery-based approach used for TRuST, and candidates without a sufficiently challenging retrieval neighborhood are removed. Runs of a strong search agent provide evidence-recall scores used to divide questions into three difficulty levels. The final selection balances search mechanic, chain depth, and difficulty, with extra emphasis on harder questions. These levels are relative to the agent used for construction, not universal difficulty labels.

\paragraph{Separation from training.}
MinHash \citep{broder1997resemblance} compares benchmark question texts with the SFT and RL question pools to remove identical and near-duplicate formulations. Additional overlap checks exclude tasks built on the same combined support sets. The text and support checks target different leakage paths: rewording a question does not necessarily change its underlying task. This separation does not remove the shared-factory relationship. Appendix~\ref{app:diagnostics} therefore also summarizes the results excluding SynthComp-En and SynthComp-Ru.

\paragraph{Evidence versus answer-bearing chunks.}
The reference evidence set contains all required chunks, including the gold subset in which the answer itself occurs. Finding that subset alone is insufficient to solve the question: the remaining supports establish constraints and connections that identify the answer. Consequently, headline SynthComp recall uses the entire evidence set, not just the answer-bearing subset.

\subsection{Scoring and evaluation protocol}
\paragraph{Recall and the output budget.}
Let $C_{10}(q)$ be the final ranked list of at most ten retrieved items for question $q$. Let $\phi_b$ map a returned identifier to the scoring unit of benchmark $b$. BrowseComp-Plus uses reference document identifiers, with $G_{q,b}$ defined by its evidence annotations. For TRuST, $\phi_b$ removes the terminal chunk suffix from a \texttt{docid\_chunkid} identifier to recover its parent document. For our SealQA adaptation and SynthComp, it preserves the evidence-chunk identifier. The selected set is
\[
S_{q,b}=\{\phi_b(c):c\in C_{10}(q)\},
\]
and recall is
\begin{equation}
R_b(q)=\frac{|S_{q,b}\cap G_{q,b}|}{|G_{q,b}|},
\end{equation}
where $G_{q,b}$ contains reference documents for BrowseComp-Plus and TRuST, annotated evidence chunks for our SealQA adaptation, and all reference evidence chunks for SynthComp. BrowseComp-Plus is scored at document level in both languages. Multiple returned chunks from the same reference document receive credit only once. The ten-item limit is applied before mapping and deduplication, so the scored set may contain fewer than ten documents. TRuST recall is not generally binary.

For SynthComp, the denominator includes all evidence chunks, not only the answer-bearing gold subset. For $N{=}3$, RRF combines three ranked lists, each containing at most ten items, and the fused ranking is truncated to ten before applying $\phi_b$ and deduplicating. Recall is averaged over questions for each benchmark. Avg.\ is the arithmetic mean of the corresponding table columns.

\paragraph{Interpreting document deduplication.}
As a hypothetical example, returned chunks \texttt{A\_1}, \texttt{A\_2}, and \texttt{B\_1} map to documents $\{A,B\}$. Against reference documents $\{A,B,C\}$, TRuST recall is $2/3$: the second chunk from $A$ gives no additional document credit.

\paragraph{Comparison scope.}
Models in the T-Search block share its memory and stopping rules. The remaining blocks evaluate adapted released search-agent harnesses over the same fixed indices, as detailed in Appendix~\ref{app:search-agent-eval}. TRuST and SynthComp are released with fixed indices.

\section{Evaluation of search-agent baselines}
\label{app:search-agent-eval}

\paragraph{Shared retrieval environment.}
For the main comparison in Table~\ref{tab:main}, all agents  search the same fixed benchmark corpora using Qwen3-Embedding-8B, without reranking or public-web access. Adapters preserve the indexed retrieval units. The text snippet for each search result is capped at 512 tokens, excluding titles and tool-response formatting. For agents supporting visit operations, returned identifiers resolve to local corpus records rather than external URLs. We do not add citation requirements or final-source-selection instructions to generative agents. Each baseline uses one rollout per question.

\paragraph{Implementation provenance.}
Our Alibaba adapters derive from the released DeepResearch code at revision \texttt{f72f75d8c3eb}, including the WebDancer and WebSailor implementations under \texttt{WebAgent}.\footnote{
\url{https://github.com/Alibaba-NLP/DeepResearch/tree/f72f75d8c3eb842f2bbbab096a12206ff66e270f}}
Harness-1 uses revision \texttt{8ac401216785},\footnote{
\url{https://github.com/pat-jj/harness-1/tree/8ac4012167858f6478fb2a8fd840e4550e2af161}}
and Search-R1 uses revision \texttt{598e61bd1d36}.\footnote{
\url{https://github.com/PeterGriffinJin/Search-R1/tree/598e61bd1d36895726d28a8d06b3a15bed19f5d3}}
We retain model-specific interaction protocols and stopping logic, subject to the adaptations below.
These experiments evaluate the released agents in our retrieval environment; they do not reproduce the authors' original benchmark settings.

\paragraph{WebDancer and WebShaper.}
WebDancer uses the released Qwen-Agent-based search-and-visit loop, with a maximum of 50 model calls and ten search results per query. For WebShaper, we adopt the same WebDancer inference recipe, as we did not identify a separate released WebShaper runner.
Both configurations retain Qwen-Agent's approximate 58k-token input limit and a maximum of 32,768 output tokens per call.
We serve both models with a 131,072-token context and fourfold YaRN \citep{peng2024yarn} scaling to accommodate these limits.

\paragraph{WebSailor.}
WebSailor uses its released tagged-text search-and-visit loop, with a 40-call budget and a native 32,768-token context without YaRN. We retain its context-limit finalization procedure which may add a final generation.

\paragraph{Tongyi DeepResearch.}
We use the released ReAct inference configuration. The available tools are local search and visit; Python execution, scholarly search, and file parsing are not exposed. The budgets are 100 model calls, 150 minutes per episode, and at most 10k output tokens per call. The model uses its native 131,072-token context without YaRN and retains context-triggered finalization.

\paragraph{Shared visit summarizer.}
Following the WebSailor setup, we use Qwen2.5-72B-Instruct for goal-conditioned extraction from visited corpus records.
All four Alibaba agents access the same auxiliary-model service. The helper uses a native 32,768-token context, with at most 24k input tokens including the visit goal and prompt formatting, and 8,192 output tokens.
Long records are truncated to fit the input budget. Decoding uses temperature zero and top-$p$ one. 
T-Search uses no auxiliary language model.

\paragraph{Harness-1.}
We retain the released Harmony interaction protocol, stateful evidence curation, and a 40-step runtime limit. Verification uses the author's unavailable-tool ablation: the verifier remains in the tool schema but returns an unavailable message without invoking a verifier model. Reranking and the downstream answer generator are disabled. The internal curated set retains its 30-item cap; our final export selects at most ten items by importance, breaking ties by insertion order.
For long-record reads, the adapter supplies a prefix of at most 4,096 tokens including a truncation marker. This avoids the upstream behavior of dropping an oversized record entirely. Like T-Search, the evaluated Harness-1 configuration uses no auxiliary language model.

\paragraph{Search-R1.}
We evaluate \path{PeterJinGo/SearchR1-nq_hotpotqa_train-qwen2.5-32b-em-grpo-v0.3} using the released search loop adapted to a completions endpoint.
It permits four search-capable generations followed by a final generation, uses greedy decoding with at most 500 output tokens per call, and maintains a rolling 4,096-token input. Each search returns three results, and the complete search observation is additionally truncated to 500 tokens. Search-R1 uses neither visit nor an auxiliary model.

\paragraph{Trajectory recall.}
For question $q$, let $H_{q,t}$ be the identifiers returned by the search backend at retrieval event $t$. Let $\phi_b$ map these identifiers to the scoring units of benchmark $b$, and define
\[
U_q = \bigcup_t \{\phi_b(h):h\in H_{q,t}\}.
\]
The mapping $\phi_b$ is the same as for final-set recall in Appendix~\ref{app:benchmarks}: BrowseComp-Plus uses document identifiers, SealQA and SynthComp use evidence-chunk identifiers, and TRuST identifiers are mapped to parent documents before deduplication. These scoring units apply to both language versions where available. Given the corresponding reference set $G_q$, trajectory recall is
\[
\mathrm{TrajectoryRecall}
=
\frac{1}{|Q|}
\sum_{q\in Q}
\frac{|U_q\cap G_q|}{|G_q|}.
\]
This metric uses all unique search-backend results, without a ten-result cutoff, rather than sources cited or selected in the final answer. For Search-R1, this includes backend-returned hits that may be truncated out of the observation shown to the policy.

T-Search and Harness-1 instead use their final selected lists of at most ten items, with the cutoff applied before benchmark-specific document mapping and deduplication. Budget-limited rollouts remain in the evaluation, and empty retrieval sets receive zero recall. Infrastructure failures are checked separately rather than silently scored as model failures.

\paragraph{Scope of comparison.}
The shared corpus, retriever, and snippet cap standardize the retrieval environment. However, agents differ in interaction budgets, context management, auxiliary-model use, and the number of sources included in their reported metric. Trajectory recall measures accumulated search coverage; final-set Recall@10 additionally measures source selection under a ten-item limit. 

\section{Additional retrieval diagnostics}
\label{app:diagnostics}

\begin{table*}[t]
\centering\small
\begin{tabular}{@{}lrrrrrrrr@{}}
\toprule
& \multicolumn{2}{c}{BrowseComp-Plus} & \multicolumn{2}{c}{SealQA} & \multicolumn{2}{c}{SynthComp} & & \\
Backend & En & Ru & En & Ru & En & Ru & TRuST & Avg. \\
\midrule
Qwen3-8B + LLM reranking & 75.04 & 66.24 & 64.82 & 59.95 & 62.71 & 62.39 & 48.93 & 62.87 \\
Qwen3-8B (default) & 65.35 & 55.95 & 61.16 & 57.72 & 54.52 & 53.13 & 43.92 & 55.96 \\
Jina v5 text-small & 60.52 & 51.41 & 62.46 & 55.60 & 56.37 & 54.56 & 39.31 & 54.32 \\
BM25 & 39.49 & 31.97 & 55.33 & 50.00 & 66.87 & 65.15 & 49.18 & 51.14 \\
Qwen3-0.6B & 51.70 & 43.71 & 56.80 & 49.53 & 54.72 & 52.27 & 36.95 & 49.38 \\
\bottomrule
\end{tabular}
\caption{Reported Recall@10 (\%) under backend substitution, with the same T-Search agent and $N{=}1$. Qwen3-8B and Qwen3-0.6B denote Qwen3-Embedding models; Jina denotes jina-embeddings-v5-text-small-retrieval \citep{akram2026jinaembeddingsv5texttasktargetedembeddingdistillation}.}
\label{tab:backends}
\end{table*}

\paragraph{Training-stage comparison.}
Table~\ref{tab:training_stages} compares the base model, the model after SFT, and the released model after SFT and RL, evaluated in the T-Search harness with Qwen3-Embedding-8B. On every benchmark, the SFT model scores above the base model, and the released model scores above the SFT model. Each trained model includes the merge of its English and Russian branches, so the comparison does not measure training and merging separately.

\begin{table}[t]
\centering
\small
\setlength{\tabcolsep}{4pt}
\begin{tabular}{lrrr}
\toprule
Benchmark & Base & SFT & +RL \\
\midrule
BrowseComp-Plus (En) & 43.7 & 54.0 & 65.4 \\
BrowseComp-Plus (Ru) & 38.6 & 47.7 & 56.0 \\
SealQA (En)          & 46.1 & 51.5 & 61.2 \\
SealQA (Ru)          & 43.3 & 50.8 & 57.7 \\
SynthComp-En         & 41.8 & 50.0 & 54.5 \\
SynthComp-Ru         & 43.9 & 48.9 & 53.1 \\
TRuST                & 33.5 & 40.2 & 43.9 \\
\midrule
Avg.                 & 41.5 & 49.0 & 56.0 \\
\bottomrule
\end{tabular}
\caption{Final-set Recall@10 (\%). Base is Qwen3.6-35B-A3B, SFT is the model after supervised fine-tuning, and +RL is the released model after SFT followed by reinforcement learning. Avg. is the unweighted mean across the benchmark rows. Scores and averages retain the one-decimal precision of the published stage comparison. The Base and +RL columns report the same results as the base model and T-Search ($N{=}1$) in Table~\ref{tab:main}, rounded to one decimal place.}
\label{tab:training_stages}
\end{table}

\paragraph{Backend substitution.}
Table~\ref{tab:backends} keeps the T-Search agent fixed while changing the search backend. Qwen3-Embedding-8B with Qwen3.6-27B as the LLM reranker gives the highest macro-average. BM25 is strongest on SynthComp and TRuST among the tested backends, while its BrowseComp-Plus scores are lower. These comparisons measure sensitivity to backend choice; they do not isolate the contribution of the multi-retriever SFT subset.

\paragraph{Results excluding SynthComp.}
To check whether the average ranking depends on the benchmark generated with the training-data factory, we recompute the unweighted mean of the five non-SynthComp columns in Table~\ref{tab:main}. Single-rollout T-Search averages 56.82 Recall@10, compared with 54.86 for GLM-5.1, the highest baseline average over these columns. These are derived averages of the reported results, not additional evaluation runs.

\paragraph{Latency study.}
Latency is measured on five randomly selected questions from each benchmark column in Table~\ref{tab:main}, 35 in total. Each question is run 13 times per configuration, with three warm-ups and ten measured repetitions. Per-query wall-clock times cover the complete agent loop and are averaged first over repetitions, then over questions within each benchmark, and finally across benchmarks with equal weights. We vary the maximum number of rounds. For $N{=}3$, the three rollouts run in parallel, so total wall-clock time includes the slowest rollout and fusion overhead. Parallel execution still requires additional inference work. Recall is evaluated on the full benchmarks, not only the 35 timing questions.

\section{Search harness implementation details}
\label{app:harness}

\paragraph{Tool contracts.}
Appendix~\ref{app:tool_schemas} presents the released function schemas as field tables and their original descriptions. The following paragraphs describe runtime validation and state transitions. The harness returns error feedback for invalid tool calls and rejects plain-text replies when a tool action is required. Validation checks tool names, arguments, and transition conditions. During SFT such invalid actions are masked from the loss while the continuation is kept (Section~\ref{sec:training}). The default per-round budget is 32,768 tokens, and the default round cap is five. The latency study varies the round cap while retaining the same transition rules.

\paragraph{Transition and stopping constraints.}
At 75\% context utilization, further search calls are locked, leaving continuation or finalization. A \texttt{save\_and\_advance} transition requires at least five searches and at least one saved chunk with an explanation. The five-search condition governs a round transition, not all final answers: a task solved earlier can finalize without making five queries. Saving is disabled in the last allowed round.

For saving and finalization, the model supplies lists of covered and unresolved question parts. Outside the final round, finalization is rejected if at least half remain unresolved. This checks the submitted coverage state; it does not independently verify the truth of the model's coverage assessment. In the last round, the agent must return its best available evidence even if the question remains partly unresolved.

\paragraph{Repeated queries and results.}
The harness rejects repeated queries differing only in case or whitespace. It also filters results already seen in the current round and chunks already carried in the context. A chunk seen in a previous round but not preserved may be retrieved again: the earlier decision to discard it does not permanently exclude it. These rules suppress unproductive repetition while allowing the agent to revisit earlier evidence after its search direction changes.

\paragraph{Final export and fallback.}
The released harness truncates the exported ranking to \texttt{max\_results}, which defaults to ten. If no \texttt{finalize\_ranking} call is accepted, it starts from the persistent saved set in insertion order and fills remaining positions with globally seen chunks ordered by cached retrieval score. The fallback can therefore include chunks from earlier rounds that were not preserved in the current context.

\subsection{Released tool schemas}
\label{app:tool_schemas}
Tables~\ref{tab:schema-search-corpus}--\ref{tab:schema-finalize-ranking} present the function schemas from the released harness at revision \texttt{997a0ba1685d}.\footnote{\href{https://github.com/turbo-llm/t-search-harness/tree/997a0ba1685d24ad840e3e2542b59952ff3fb362}{Released harness source, revision \texttt{997a0ba1685d}.}}
They preserve field names, types, required-field membership, defaults, and the original descriptions. For nested fields, Required refers to the enclosing object; \texttt{[]} denotes an array item. Empty description cells correspond to fields without a description in the source schema. Descriptions in the tables are copied from the schemas, with only LaTeX typesetting applied. Before sending the schemas to the model, the harness replaces configuration placeholders in the descriptions with their configured values. With the release defaults, \texttt{\_\_MIN\_SEARCHES\_BEFORE\_SAVE\_\_} is replaced by 5 searches, and the minimum lengths for \texttt{reason}, \texttt{round\_summary}, and \texttt{next\_round\_goal} are 20, 50, and 30 characters, respectively. The harness also validates tool calls against the transition and stopping rules described above.

\begin{table*}[tp]

\normalsize

\noindent\begin{tabularx}{\textwidth}{@{}>{\raggedright\arraybackslash}p{0.28\textwidth}>{\raggedright\arraybackslash}p{0.14\textwidth}l>{\raggedright\arraybackslash}X@{}}

\toprule

\multicolumn{4}{@{}p{\textwidth}@{}}{\textbf{\texttt{search\_\allowbreak{}corpus}}} \\ \addlinespace[4pt]

\multicolumn{4}{@{}p{\textwidth}@{}}{Semantic search over the corpus. Returns up to top\_\allowbreak{}k items as \{chunk\_\allowbreak{}id, snippet, score\}. Parallel calls allowed --- issue multiple search\_\allowbreak{}corpus tool calls in one assistant turn to cover independent angles. Dedup: within the current round, chunks already shown this round are filtered. Chunks in your saved set are filtered (already in context at top). Chunks seen in PREVIOUS rounds but not saved CAN resurface --- re-consider them under the new angle.} \\ \midrule

Field & Type & Required & Description \\

\midrule

\path{query} & string & Yes & Search query string. \\ \addlinespace[3pt]

\path{top_k} & integer & No & Number of results to return. \textit{Default: 5.} \\ \addlinespace[3pt]

\bottomrule

\end{tabularx}

\caption{Released schema for \texttt{search\_\allowbreak{}corpus}. Tool type: \texttt{function}; parameters: \texttt{object}.}

\label{tab:schema-search-corpus}

\end{table*}

\begin{table*}[tp]

\normalsize

\noindent\begin{tabularx}{\textwidth}{@{}>{\raggedright\arraybackslash}p{0.28\textwidth}>{\raggedright\arraybackslash}p{0.14\textwidth}l>{\raggedright\arraybackslash}X@{}}

\toprule

\multicolumn{4}{@{}p{\textwidth}@{}}{\textbf{\texttt{save\_\allowbreak{}and\_\allowbreak{}advance}}} \\ \addlinespace[4pt]

\multicolumn{4}{@{}p{\textwidth}@{}}{End the current round and start a new one with fresh context. Only saved chunks survive the transition. Requires $\geq$\_\allowbreak{}\_\allowbreak{}MIN\_\allowbreak{}SEARCHES\_\allowbreak{}BEFORE\_\allowbreak{}SAVE\_\allowbreak{}\_\allowbreak{} searches in the current round and $\geq$1 chunk saved. DISABLED in the last round. chunk\_\allowbreak{}ids: saved come from this round's search results or the currently-saved set (re-passing updates the reason); drop from the current saved set; no overlap between saved\_\allowbreak{}chunks and drop\_\allowbreak{}from\_\allowbreak{}saved. You MUST declare progress via covered\_\allowbreak{}concepts and unresolved\_\allowbreak{}concepts --- at least one of the two must be non-empty, and they must not overlap.} \\ \midrule

Field & Type & Required & Description \\

\midrule

\path{saved_chunks} & array of object & Yes & Chunks to append to / update in the saved set. \\ \addlinespace[3pt]

\path{saved_chunks[].chunk_id} & string & Yes &  \\ \addlinespace[3pt]

\path{saved_chunks[].reason} & string & Yes & $\geq$\_\allowbreak{}\_\allowbreak{}MIN\_\allowbreak{}REASON\_\allowbreak{}LEN\_\allowbreak{}\_\allowbreak{} chars. Specific fact or entity this chunk contributes. \\ \addlinespace[3pt]

\path{drop_from_saved} & array of string & No & Optional. chunk\_\allowbreak{}ids to remove from the saved set. Use [] if dropping nothing. \\ \addlinespace[3pt]

\path{covered_concepts} & array of string & Yes & Query atoms now supported by saved chunks. Short labels (entities, dates, attributes). Can be [] if no solid evidence yet, but then unresolved\_\allowbreak{}concepts must be non-empty. \\ \addlinespace[3pt]

\path{unresolved_concepts} & array of string & Yes & Query atoms still without evidence. Short labels. Can be [] only if all concepts are covered (then covered\_\allowbreak{}concepts must list them). \\ \addlinespace[3pt]

\path{round_summary} & string & Yes & $\geq$\_\allowbreak{}\_\allowbreak{}MIN\_\allowbreak{}ROUND\_\allowbreak{}SUMMARY\_\allowbreak{}LEN\_\allowbreak{}\_\allowbreak{} chars. What you searched this round and what you found. \\ \addlinespace[3pt]

\path{next_round_goal} & string & Yes & $\geq$\_\allowbreak{}\_\allowbreak{}MIN\_\allowbreak{}NEXT\_\allowbreak{}GOAL\_\allowbreak{}LEN\_\allowbreak{}\_\allowbreak{} chars. Concrete plan targeting unresolved. \\ \addlinespace[3pt]

\bottomrule

\end{tabularx}

\caption{Released schema for \texttt{save\_\allowbreak{}and\_\allowbreak{}advance}. Tool type: \texttt{function}; parameters: \texttt{object}.}

\label{tab:schema-save-and-advance}

\end{table*}

\begin{table*}[tp]

\normalsize

\noindent\begin{tabularx}{\textwidth}{@{}>{\raggedright\arraybackslash}p{0.28\textwidth}>{\raggedright\arraybackslash}p{0.14\textwidth}l>{\raggedright\arraybackslash}X@{}}

\toprule

\multicolumn{4}{@{}p{\textwidth}@{}}{\textbf{\texttt{finalize\_\allowbreak{}ranking}}} \\ \addlinespace[4pt]

\multicolumn{4}{@{}p{\textwidth}@{}}{Submit final ranked list of most relevant documents for the user query and end the session. Empty ranking rejected. chunk\_\allowbreak{}ids must come from your current context (saved + this round's seen). You MUST declare covered\_\allowbreak{}concepts and unresolved\_\allowbreak{}concepts --- harness will reject finalize if unresolved $\geq$ 50\% of (covered + unresolved), except in the last round (where save\_\allowbreak{}and\_\allowbreak{}advance is disabled). Retry allowed on reject.} \\ \midrule

Field & Type & Required & Description \\

\midrule

\path{ranking} & array of object & Yes & Ordered most$\rightarrow$least relevant. \\ \addlinespace[3pt]

\path{ranking[].chunk_id} & string & Yes &  \\ \addlinespace[3pt]

\path{ranking[].reason} & string & Yes & $\geq$\_\allowbreak{}\_\allowbreak{}MIN\_\allowbreak{}REASON\_\allowbreak{}LEN\_\allowbreak{}\_\allowbreak{} chars. Specific fact or entity this chunk contributes. \\ \addlinespace[3pt]

\path{covered_concepts} & array of string & Yes & Query atoms supported by evidence in the final ranking. Short labels. \\ \addlinespace[3pt]

\path{unresolved_concepts} & array of string & Yes & Query atoms still without evidence at finalize time. Short labels. \\ \addlinespace[3pt]

\bottomrule

\end{tabularx}

\caption{Released schema for \texttt{finalize\_\allowbreak{}ranking}. Tool type: \texttt{function}; parameters: \texttt{object}.}

\label{tab:schema-finalize-ranking}

\end{table*}

\subsection{System prompt and round re-entry}
\label{app:agent_prompts}
The system-prompt template in Table~\ref{tab:system-prompt} is reproduced from the same release. The harness replaces \texttt{\_\_MAX\_ROUNDS\_\_} with the configured round cap. Round 1 pairs this system message with a user message of the form \texttt{Query: \{query\}}. Each subsequent round uses the same system message and the re-entry template in Table~\ref{tab:round-reentry}. Re-entry carries saved evidence, round summaries, coverage declarations, and the next-round goal, rather than the full preceding transcript.

\begin{table*}[tp]
\begin{lstlisting}[style=tsearchprotocol]
You are a document retrieval specialist. First stage of a two-stage pipeline:

    user query (*@\(\rightarrow\)@*) YOU (*@\(\rightarrow\)@*) ranked documents (*@\(\rightarrow\)@*) downstream agent (*@\(\rightarrow\)@*) answer

You do NOT answer the question. You output a ranked list of documents via `finalize_ranking` (*@\textemdash{}@*) the only normal way to end the session.

# Rounds

You operate in up to __MAX_ROUNDS__ rounds (a ceiling, not a target). Each round is a fresh 32K-token context for searching the corpus. Chunks saved via `save_and_advance` are the only state that carries across rounds. A round ends with `save_and_advance` (continue in a new round) or `finalize_ranking` (end the session). **Call `finalize_ranking` from any round the moment evidence is sufficient.**

At 75% of round budget `search_corpus` locks (*@\textemdash{}@*) you must then `save_and_advance` or `finalize_ranking`. In the last round only `finalize_ranking` is available. `finalize_ranking` is rejected when (*@\(\geq\)@*)50% of declared concepts are unresolved and you aren't in the last round (*@\textemdash{}@*) use `save_and_advance` instead.

Each tool response carries:

    [Round N/__MAX_ROUNDS__ | Budget: X/32768 (pct%) | Saved: K chunks | Searches this round: M]

# Available tools

- `search_corpus` (*@\textemdash{}@*) semantic corpus search within the current round.
- `save_and_advance` (*@\textemdash{}@*) end the round, carry the saved set to the next.
- `finalize_ranking` (*@\textemdash{}@*) submit the final ranked list and end the session.

# Strategy

Identify the query's key concepts. Search from distinct angles, follow up on entities and bridges you find. Finalize honestly when the corpus stops yielding new evidence.

# Hard rules

Every turn has (*@\(\geq\)@*)1 tool call (*@\textemdash{}@*) no free-text replies. Follow explicit textual evidence, not speculation.

# Output format (via finalize_ranking)

```
{
  "ranking": [{"chunk_id": "...", "reason": "..."}, ...],
  "covered_concepts": [...],
  "unresolved_concepts": [...]
}
```
\end{lstlisting}\caption{System-prompt template from the pinned harness release.}\label{tab:system-prompt}\end{table*}

\paragraph{Round re-entry template.}
The brace-delimited fields are filled from the current query and session state. The saved block contains chunk identifiers, reasons, and snippets. The history block contains previous round goals, summaries, and coverage declarations. The coverage state aggregates declared covered and unresolved concepts. The \texttt{\{save\_tool\_note\}} field contains the corresponding note in Table~\ref{tab:save-notes} depending on whether another round is available.

\begin{table*}[tp]
\begin{lstlisting}[style=tsearchprotocol]
# Original query

{original_query}

# Saved chunks (carried over)

{saved_block}

# Round history

{history_block}

# Coverage state

{coverage_state}

# Round {current_round} of {max_rounds}

## Goal set in round {prev_round}:
{round_goal}

# Task

Re-read saved chunks, coverage state, and the goal. Decide:

1. **Is the goal still valid?** Adjust if new findings shifted priorities.
2. **Do saved chunks already cover most of the query?** If yes, `finalize_ranking` now (*@\textemdash{}@*) don't burn a round for marginal gain.
3. **Otherwise pursue unresolved concepts** with fresh angles. If a concept resisted 2+ rounds, evidence is likely not in the corpus (*@\textemdash{}@*) finalize honestly with it declared unresolved.

Reminder: {max_rounds} is a ceiling, not a target.

{save_tool_note}

Budget: 32K this round. At 75% search locks (*@\textemdash{}@*) `save_and_advance` or `finalize_ranking`.
\end{lstlisting}
\caption{Round re-entry template. Brace-delimited fields are filled from the current query and session state.}\label{tab:round-reentry}\end{table*}

\begin{table*}[tp]\noindent\begin{minipage}[t]{0.48\textwidth}\textbf{Before the last round}\par
\begin{lstlisting}[style=tsearchprotocol]
save_and_advance is available after 5+ searches this round. Use it only when fresh context would genuinely help close remaining concepts (*@\textemdash{}@*) if another round is unlikely to add evidence, finalize_ranking is the right call.
\end{lstlisting}
\end{minipage}\hfill%
\begin{minipage}[t]{0.48\textwidth}\textbf{In the last round}\par
\begin{lstlisting}[style=tsearchprotocol]
Last round (__MAX_ROUNDS__). save_and_advance is DISABLED (*@\textemdash{}@*) call finalize_ranking before the budget runs out. The harness does NOT reject finalize here even with many unresolved concepts (*@\textemdash{}@*) submit your best candidates with honest covered/unresolved declarations.
\end{lstlisting}
\end{minipage}\caption{Alternative values of the \texttt{\{save\_tool\_note\}} field.}\label{tab:save-notes}\end{table*}

\end{document}